\documentclass{article}
\usepackage{iclr2021_conference,times}
\usepackage{amsmath}
\usepackage{amssymb}
\usepackage{hyperref}
\usepackage{url}

\usepackage[pdftex]{graphicx} 
\usepackage{caption}   

\usepackage{enumitem}
\usepackage{booktabs}
\usepackage{xspace}
\usepackage{xcolor}
\usepackage{wrapfig}
\usepackage{xfrac}
\usepackage{tabularx}
\usepackage{tikz}

\definecolor{natural}{rgb}{0.7137,0.3333,0.3333}
\definecolor{specialized}{rgb}{0.4118,0.6431,0.4314}
\definecolor{structured}{rgb}{0.3254,0.4431,0.6666}
\definecolor{all}{rgb}{0.7529,0.4902,0.6471}

\definecolor{alexey}{rgb}{0.8, 0.0, 0.8}

\definecolor{matthias}{rgb}{0.0, 0.8, 0.8}

\definecolor{sylvain}{rgb}{0.8, 0.8, 0.0}

\usepackage{fancyhdr}
\fancypagestyle{plain}{
    \fancyhf{} 
}
\title{\centering Pixels to Signals:\\A Real-Time Framework for Traffic Demand Estimation}

\author{
\centerline{Hrithik Mhatre*, Mohak Vyas*, Archak Mittal*$^{\dagger}$} \vspace{0.5mm} \\
\centerline{$^\dagger$Corresponding Author and Assistant Professor}  \vspace{0.7mm} \\
\centerline{* Civil Engineering Department, Indian Institute of Technology Bombay}\\
\centerline{\texttt{archak@iitb.ac.in}} \\
}

\iclrfinalcopy %
\begin{document}

\maketitle
\begin{abstract}
Traffic congestion is becoming a challenge in the rapidly growing urban cities, resulting in increasing delays and inefficiencies within urban transportation systems. To address this issue a comprehensive methodology is designed to optimize traffic flow and minimize delays. The framework is structured with three primary components: (a) vehicle detection, (b) traffic prediction, and (c) traffic signal optimization. This paper presents the first component, vehicle detection. The methodology involves analyzing multiple sequential frames from a camera feed to compute the background, i.e. the underlying roadway, by averaging pixel values over time. The computed background is then utilized to extract the foreground, where the Density-Based Spatial Clustering of Applications with Noise (DBSCAN) algorithm is applied to detect vehicles. With its computational efficiency and minimal infrastructure modification requirements, the proposed methodology offers a practical and scalable solution for real-world deployment.
\end{abstract}

\section{Introduction}

Rapid urbanization of the modern world has intensified the demand for efficient and intelligent transportation systems. Traditional traffic management approaches, primarily driven by static and heuristic-based methods, often fail to adapt to the dynamic and complex nature of urban traffic. 
Moreover, the current intelligent transportation systems are not adapted to the complexities and chaos of a developing country. This leads to congestion, increased environmental pollution, and reduced economic efficiency.

This paper explores an approach to optimizing traffic signals across a bounded urban network using an integrated framework of computer vision for real-time traffic demand estimation, machine learning for traffic demand prediction, and reinforced learning for network optimization. 

The proposed solution consists of three main components :
\vspace{-4pt}
\begin{enumerate} 
    \item Detection: Using computer vision techniques to detect vehicles in real time.
\vspace{-4pt}
    \item Prediction: Using machine learning to predict traffic demand using spatial and temporal data.
\vspace{-4pt}
    \item Optimization: Employ a reinforcement learning-based framework that dynamically adjusts signal timings.
\end{enumerate}

The methodology proposed in this paper focuses primarily on the first component: real-time vehicle detection.

\newcommand{\op}[1]{\operatorname{#1}}
\newcommand{\mbf}[1]{\mathbf{#1}}

\section{Methodology}

The real-time vehicle detection component is implemented in two stages. In the first stage, the background is modeled and subtracted from the video frames to isolate the foreground, which represents moving objects or regions of interest in the scene, such as vehicles. This is followed by the application of image processing techniques to refine the foreground and minimize noise. In the second stage, the DBSCAN (Density-Based Spatial Clustering of Noise-Based Applications) \cite{dbscanArticle} algorithm detects vehicles within the processed foreground.

\subsection{Foreground Isolation and Noise Reduction}

Foreground isolation from a video becomes relatively straightforward when the background is known. However, challenges arise in scenarios where the background is unknown and must be modeled using available video data. 

\vspace{-7pt}

\paragraph{Background Reconstruction from Sampled Frames:}
Given a video feed, we sample frames at regular intervals ($\Delta t = 0.5$ seconds in this study) to construct a set $\mathcal{F} = \{f_1, f_2, \ldots, f_n\}$ of sampled frames. The underlying assumption is that while individual frames may occlude parts of the background due to transient objects, the aggregated set $\mathcal{F}$ contains sufficient information to reconstruct the complete static background. 

\begin{figure}[ht]
\begin{center}
\begin{tabular}{c}
\includegraphics[width=1.0\textwidth]{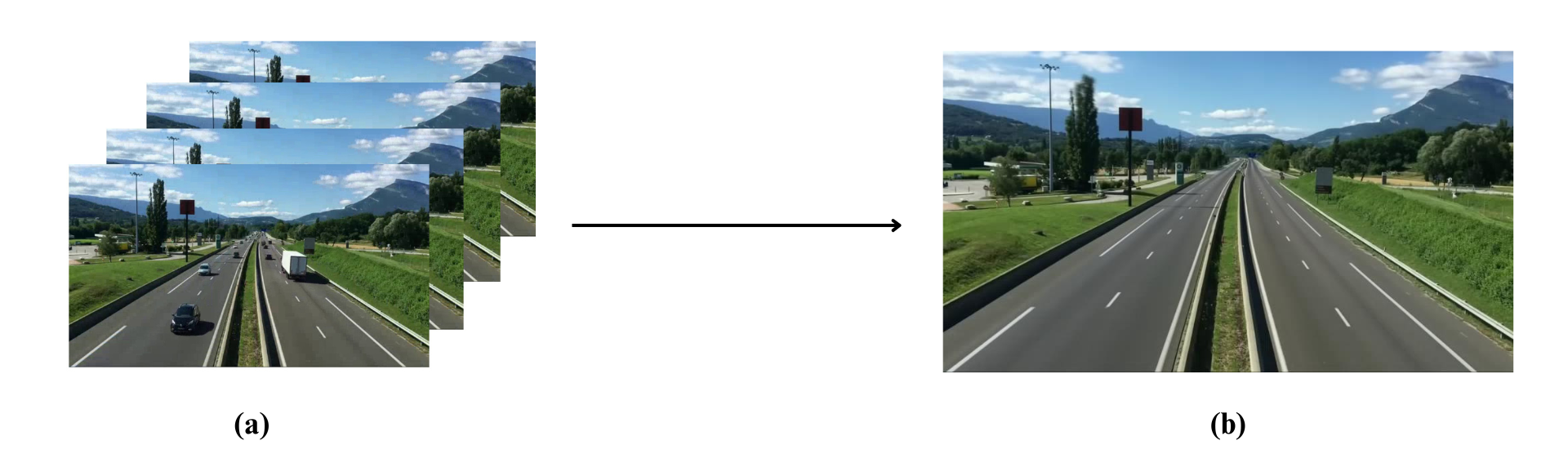}
\end{tabular}
\end{center}
\caption{Background reconstruction: (a) Sampled frames $F = \{f_1, f_2, \ldots, f_n\}$ at regular intervals $\Delta t$, and (b) the computed background $f_{bg}$ generated through pixel-wise mean of all frames. The operation $f_{bg} = \frac{1}{n}\sum_{i=1}^n f_i$ produces a static representation where transient objects are suppressed.}
\label{fig:model_background}
\end{figure}

The background frame $f_{\text{bg}}$ is computed as the pixel-wise mean of all sampled frames:

\[
f_{\text{bg}}(x,y) = \frac{1}{n} \sum_{i=1}^{n} f_i(x,y),
\]

where $f_i(x,y)$ denotes the intensity at the pixel $(x,y)$ in the $i$-th frame, and $n$ represents the total number of sampled frames. This averaging process suppresses dynamic components (which vary across frames) while reinforcing static structures, yielding a smooth and noise-free background estimate. For our analysis, approximately 670 frames were sampled at the specified interval. The operations are illustrated in Figure~\ref{fig:model_background}.

\paragraph{Foreground Isolation and Image Preprocessing:}

For each frame \( f_i \), the foreground extraction process consists of:

\[
f_{i,fg} = (\mathcal{M} \circ \mathcal{B} \circ \mathcal{G} \circ \mathcal{D})(f_i)
\]

where:
\begin{itemize}
    \item \( \mathcal{D}(f_i) = |f_i - f_{bg}| \) is the background differencing operation
    \item \( \mathcal{G} \) denotes the grayscale conversion function
    \item \( \mathcal{B} \) denotes the binarization operation with threshold \( \tau \)
    \item $\mathcal{M} = \mathcal{\oplus} \circ \mathcal{\ominus}$ (erosion $\mathcal{\ominus}$ followed by dilation $\mathcal{\oplus}$)
    \item \( \circ \) denotes function composition
\end{itemize}

\begin{figure}[ht]
\begin{center}
\begin{tabular}{c}
\includegraphics[width=1.0\textwidth]{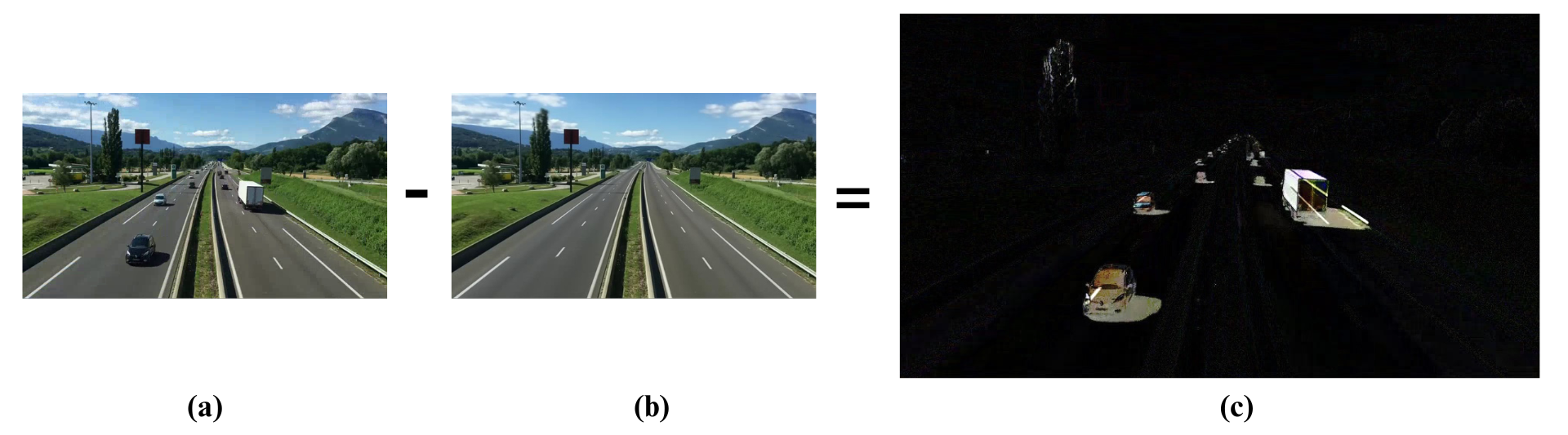}
\end{tabular}
\end{center}
\caption{Foreground isolation: (a) Original frame $f_i$; (b) Background model $f_{bg}$; (c) Isolated foreground $\mathcal{D}(f_i) = |f_i - f_{bg}|$}
\label{fig:model_foreground}
\end{figure}

\begin{figure}[ht]
\begin{center}
\begin{tabular}{c}
\includegraphics[width=1.0\textwidth]{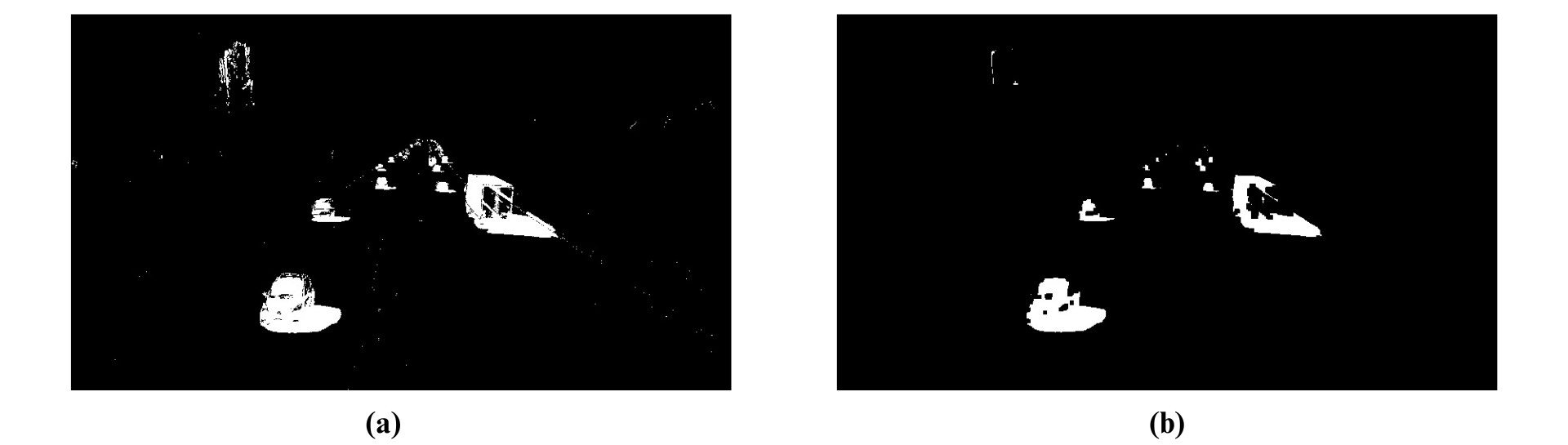}
\end{tabular}
\end{center}
\caption{Image preprocessing stages: (a) Grayscale and binarization ($\mathcal{G} \rightarrow \mathcal{B}$) to highlight potential vehicle regions (b) Morphological refinement ($\mathcal{M} = \mathcal{\oplus} \circ \mathcal{\ominus}$) to refine the regions and reduce noise}
\label{fig:model_cv}
\end{figure}

The pipeline transforms each input frame \( f_i \) into its foreground \( f_{i,fg} \), where foreground regions are highlighted and background suppressed. The operations for foreground isolation and image preprocessing are depicted in Figure \ref{fig:model_foreground} and Figure \ref{fig:model_cv}, respectively. The image processing operations of Grayscale ($\mathcal{G}$), Binarization ($\mathcal{B}$), Erosion ($\mathcal{\ominus}$), and Dilation ($\mathcal{\oplus}$) are discussed in detail in the following sections.

\label{sec:grayscaling}
\textbf{Grayscaling ($\mathcal{G}$): }
Grayscaling is the process of converting a color image (typically in RGB format) to a single-channel grayscale image where each pixel represents only the brightness information. This transformation simplifies image processing tasks while preserving the essential structural information of the image.

The standard conversion from an RGB color space to grayscale uses a weighted average of the color channels that accounts for human perceptual sensitivity to different colors. The conversion is given by:

\[
Y \leftarrow 0.299 \cdot R + 0.587 \cdot G + 0.114 \cdot B
\]

where $R$, $G$, and $B$ represent the red, green, and blue color channels respectively, and $Y$ represents the resulting grayscale intensity value. The weights (0.299, 0.587, 0.114) reflect the different luminosity perception of the human eye to each color channel, with green contributing the most and blue the least.

\label{sec:binarization}
\textbf{Binarization ($\mathcal{B}$): }
Image binarization is the process of converting a grayscale image into a binary (black-and-white) image using a threshold value. This fundamental operation in image processing serves to simplify the image data while preserving essential shape information.

Let \( T \) be the threshold (a scalar in \([0, 255]\)), then:

\[
I_{\text{binary}}(x, y) = 
\begin{cases} 
1 \ (\text{or } 255) & \text{if } I_{\text{gray}}(x, y) \geq T \\
0 & \text{otherwise}
\end{cases}
\]

This thresholding operation is mathematically equivalent to applying a Heaviside step function to each pixel:

\[
I_{\text{binary}}(x, y) = H(I_{\text{gray}}(x, y) - T)
\]

where the Heaviside function \( H \) is defined as:
\[
H(z) = 
\begin{cases} 
1 & \text{if } z \geq 0 \\
0 & \text{otherwise}
\end{cases}
\]

Here, \( I_{\text{gray}}(x, y) \) represents the grayscale intensity at pixel \((x, y)\), and \( I_{\text{binary}}(x, y) \) is the resulting binary value. The choice between using 1 or 255 for the white value depends on the specific implementation requirements.

\label{sec:erosion}
\textbf{Erosion ($\mathcal{\ominus}$): }
With \( A \) and \( B \) as sets in \( \mathbb{Z}^2 \), the erosion of \( A \) by \( B \), denoted \( A \ominus B \), is defined as

\[
A \ominus B = \{ z | (B)_z \subseteq A \} 
\]

In simple words, this equation indicates that the erosion of \( A \) by \( B \) is the set of all points \( z \) such that \( B \), translated by \( z \), is contained in \( A \). In this and the following notation we assume B as a structuring element and A is the image object on which the operations are being performed.

\label{sec:dilation}
\textbf{Dilation ($\mathcal{\oplus}$): }
With \( A \) and \( B \) as sets in \( \mathbb{Z}^2 \), the dilation of \( A \) by \( B \), denoted \( A \oplus B \), is defined as

\[
A \oplus B = \{ z | (\hat{B})_z \cap A \neq \emptyset \}.
\]

This equation is based on reflecting \( B \) about its origin, and shifting this reflection by \( z \). The dilation of \( A \) by \( B \) then is the set of all displacements, \( z \), such that \( \hat{B} \) and \( A \) overlap by at least one element.

The approach proposed in this study operates effectively under specific assumptions, which are outlined below:
\begin{enumerate}[noitemsep, topsep=0pt] 
    \item The foreground objects are in motion.
    \item The foreground objects are relatively small in comparison to the background.
    \item The video maintains consistent lighting conditions.
\end{enumerate}

\subsection{DBSCAN Algorithm}

DBSCAN \cite{dbscanArticle} is a density-based clustering algorithm. The fundamental concept of density-based clustering is to detect areas of high and low density, effectively distinguishing regions of high density from those of low density. The algorithm relies on two key parameters: $\epsilon$ (epsilon) and \textit{MinPts} (minimum samples). $\epsilon$ defines a distance threshold that determines the neighborhood around a point, where points within this radius are considered neighbors and can be grouped into clusters. \textit{MinPts} specifies the minimum number of points required to form a dense region, enabling the identification of clusters and noise.

Given a dataset $D = \{p_1, p_2, \dots, p_n\}$, the DBSCAN algorithm proceeds as follows:

\begin{table}[ht]
\centering
\begin{tabular}{>{\bfseries}l p{0.8\linewidth}}
\toprule
Step & Description \\
\midrule
Initialization & Begin with an arbitrary point $p_i$. If $p_i$ is unvisited, mark it as visited. \\
\addlinespace
Neighborhood Search & Compute the $\epsilon$-neighborhood of $p_i$: $N_\epsilon(p_i) = \{p_j \mid \text{distance}(p_i, p_j) \leq \epsilon\}$ \\
\addlinespace
Core Point Check & If $|N_\epsilon(p_i)| \geq \text{MinPts}$, then $p_i$ is classified as a core point and a new cluster is initiated. If $|N_\epsilon(p_i)| < \text{MinPts}$, $p_i$ is labeled as noise (but may later be included in a cluster if it's a border point of a core point). \\
\addlinespace
Cluster Expansion & For each core point, all points within its $\epsilon$-neighborhood are added to the current cluster. The cluster expands recursively by visiting the neighborhoods of newly added points. If any are core points, their neighborhoods are also expanded. \\
\addlinespace
Repeat & Process continues until all points are visited. Unassigned points are considered noise. The final output consists of clusters of dense points with isolated noise points. \\
\bottomrule
\end{tabular}
\end{table}

\begin{figure}[ht]
\begin{center}
\begin{tabular}{c}
\includegraphics[width=.5\textwidth]{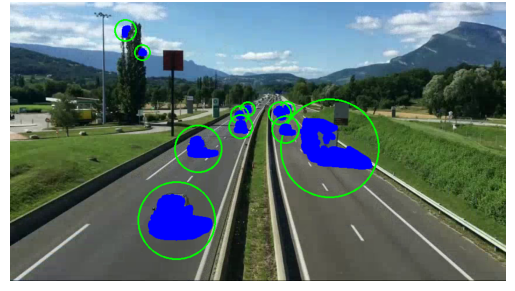}
\end{tabular}
\end{center}
\caption{Clusters are generated using the DBSCAN clustering algorithm, highlighting distinct groupings of data points corresponding to detected vehicles.}
\label{fig:model_clusters}
\end{figure}

In this study, we utilized the DBSCAN algorithm to effectively group vehicle pixels extracted from video frames. After converting the foreground to grayscale and applying morphological operations (Figure~\ref{fig:model_cv}), we clustered the detected blobs using DBSCAN. The algorithm's hyperparameters, $\epsilon$ and \textit{MinPts}, were carefully tuned to group nearby centroids corresponding to individual vehicles. Noise points (outliers) were excluded, and the number of vehicles was determined by counting the unique clusters. To visually confirm the clustering results, we highlighted the detected vehicles by drawing circles around them, as shown in Figure~\ref{fig:model_clusters}.

\section{Related Work}

\paragraph{YOLO (You Only Look Once):}
YOLO \cite{yolov5} is a real-time object detection algorithm that uses a single convolutional neural network (CNN) to detect and classify objects within an image. It divides the image into a grid and predicts bounding boxes and class probabilities for each cell. Unlike traditional object detection methods that apply classifiers to image regions in multiple passes, YOLO processes the entire image in one forward pass, enabling high-speed detection. This method excels in recognizing objects with clear features but may struggle with small or distant objects and complex backgrounds. In this study, YOLOv5 is utilized as a baseline for vehicle detection to compare its performance with the proposed clustering-based approach.

\paragraph{K-means Algorithm:}

\textit{K-means} \cite{kmeans_original} is a centroid-based clustering algorithm. The fundamental concept of K-means clustering is to partition data into $k$ distinct clusters where each observation belongs to the cluster with the nearest mean. The algorithm aims to minimize the within-cluster sum of squares (WCSS):

\[
    \text{WCSS} = \sum_{i=1}^{k} \sum_{x \in C_i} \|x - \mu_i\|^2
\]

where $C_i$ represents the $i^{th}$ cluster and $\mu_i$ is the centroid of cluster $C_i$.

The algorithm relies on one key parameter: $k$ (number of clusters). Given a dataset $D = \{x_1, x_2, \dots, x_n\}$ where each $x_i \in \mathbb{R}^d$, the K-means algorithm proceeds as follows:

\begin{table}[h]
\centering
\begin{tabular}{>{\bfseries}l p{0.8\linewidth}}
\toprule
Step & Description \\
\midrule
Initialization & Randomly select $k$ initial centroids $\{\mu_1^{(0)}, \mu_2^{(0)}, \dots, \mu_k^{(0)}\}$ from the dataset points. \\
\addlinespace
Assignment & Assign each data point $x_i$ to the nearest centroid:
\[ C_j^{(t)} = \{x_i : \|x_i - \mu_j^{(t)}\| \leq \|x_i - \mu_l^{(t)}\| \ \forall l \neq j\} \]
This creates $k$ clusters $C_1^{(t)}, C_2^{(t)}, \dots, C_k^{(t)}$ at iteration $t$. \\
\addlinespace
Update & Recalculate centroids as the mean of all points in each cluster:
\[ \mu_j^{(t+1)} = \frac{1}{|C_j^{(t)}|} \sum_{x_i \in C_j^{(t)}} x_i \] \\
\addlinespace
Convergence Check & Repeat the Assignment and Update steps until either:
\begin{itemize}
    \item The centroids stabilize ($\mu_j^{(t+1)} \approx \mu_j^{(t)}$)
    \item The maximum number of iterations is reached
    \item The WCSS improvement falls below a threshold
\end{itemize} \\
\bottomrule
\end{tabular}
\end{table}

Unlike DBSCAN, K-means requires pre-specifying the number of clusters and tends to produce convex, equally-sized clusters.

\paragraph{K-means for background reconstruction:}
The K-means based background reconstruction \cite{video_background_removal} clusters pixel-level data across video frames to differentiate between background and foreground pixels. For each pixel location, the algorithm collects intensity (or RGB) values across frames and applies K-means clustering with $k$=2. One cluster represents the relatively static background, while the other corresponds to the dynamic foreground. The background cluster, typically the one with the largest number of points, is reconstructed by calculating the median intensity or RGB value. This reconstructed background serves as a stable reference, effectively isolating the background from any foreground variations.
\section{Results}

The proposed background reconstruction and clustering method was evaluated against standard approaches, including K-means for background reconstruction and YOLOv5 for vehicle detection, with runtime performance analyzed for a more comprehensive comparison. 

\begin{figure}[h]
    \centering
    \includegraphics[width=0.83\textwidth]{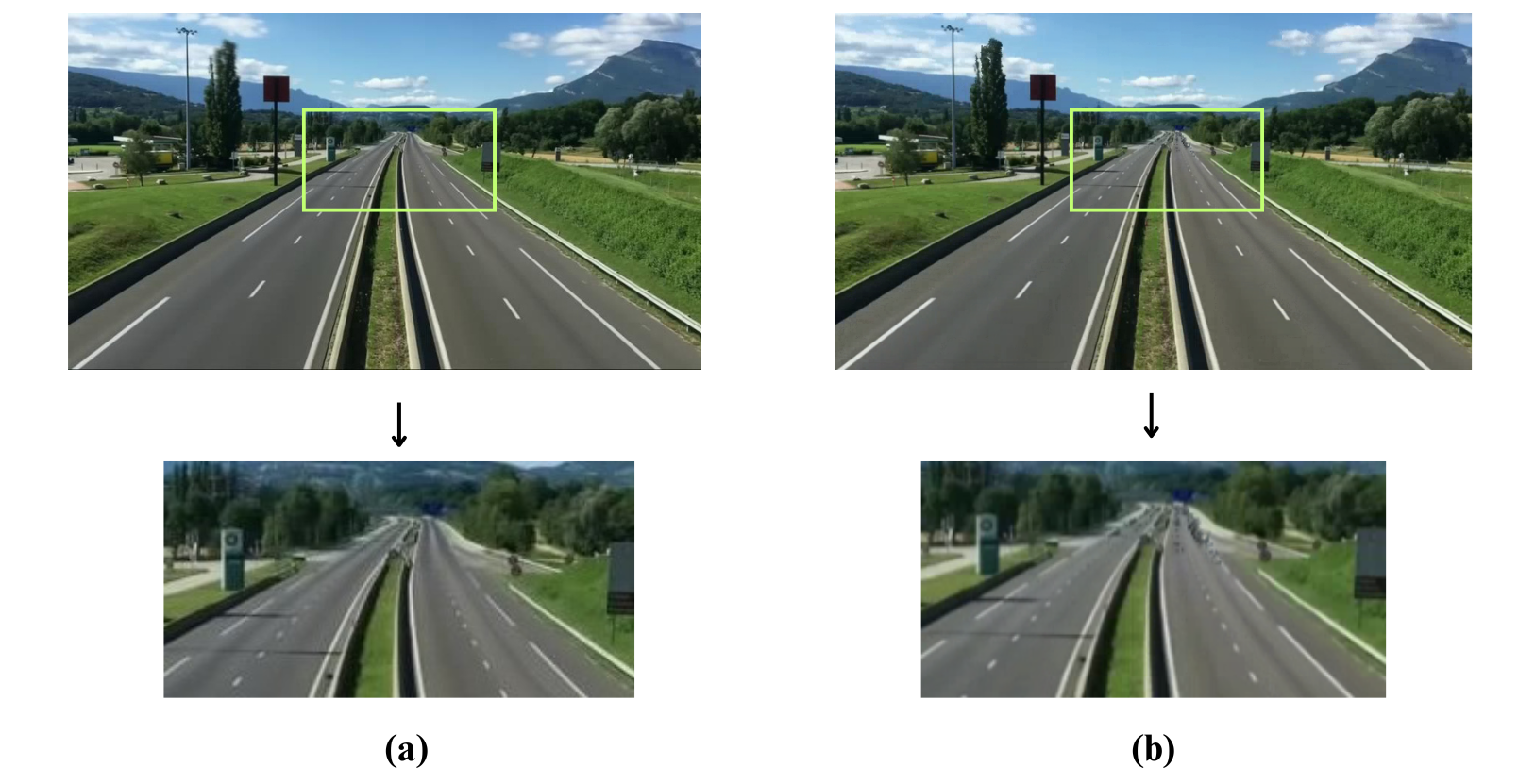}
    \caption{Comparison of extracted backgrounds: (a) Proposed method and (b) K-means. The zoomed-in view at the bottom clearly illustrates that the proposed method provides a clearer background.}
    \label{fig:background_comparison}
\end{figure}

\paragraph{Evaluation of Background Reconstruction Methods:}
The background reconstructed using the proposed method was noticeably clearer and smoother, especially at the far end of the frame, when compared to the K-means method, as illustrated in Figure~ \ref{fig:background_comparison}.

\begin{figure}[h]
    \centering
    \includegraphics[width=1\textwidth]{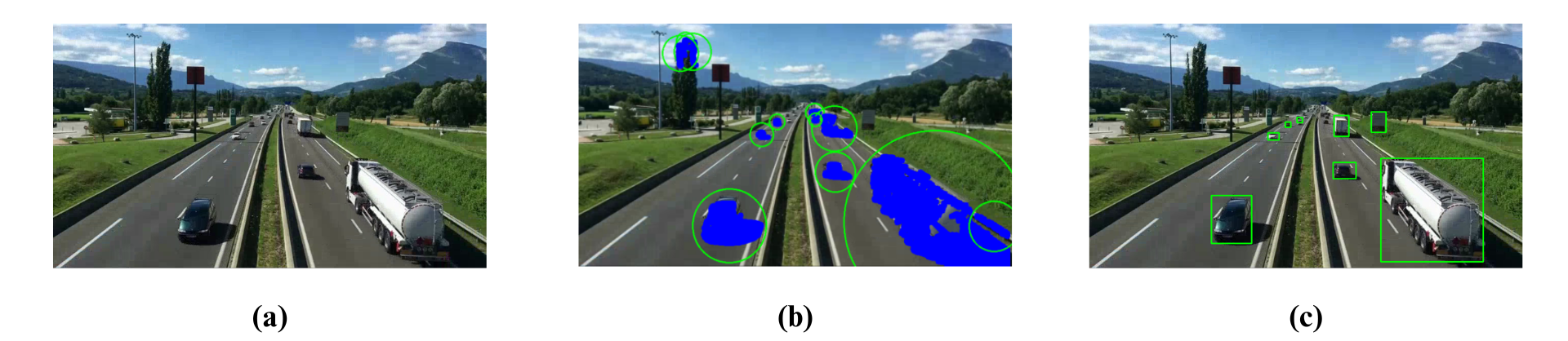}
    \caption{Vehicle detection comparison: (a) Original frame displayed for reference, (b) Detection results using DBSCAN, and (c) Detection results using YOLOv5.}
    \label{fig:detection_comparison}
    \vspace{-2mm}
\end{figure}

\paragraph{Evaluation of Detection Algorithms:}
The proposed vehicle detection algorithm, which utilizes DBSCAN clustering, demonstrated superior performance in identifying vehicles within a frame compared to YOLOv5. DBSCAN effectively detected vehicles even in distant parts of the frame, which YOLOv5 often missed. These results are evident in Figure \ref{fig:detection_comparison}

\begin{figure}[h]
    \centering
    \includegraphics[width=.4\textwidth]{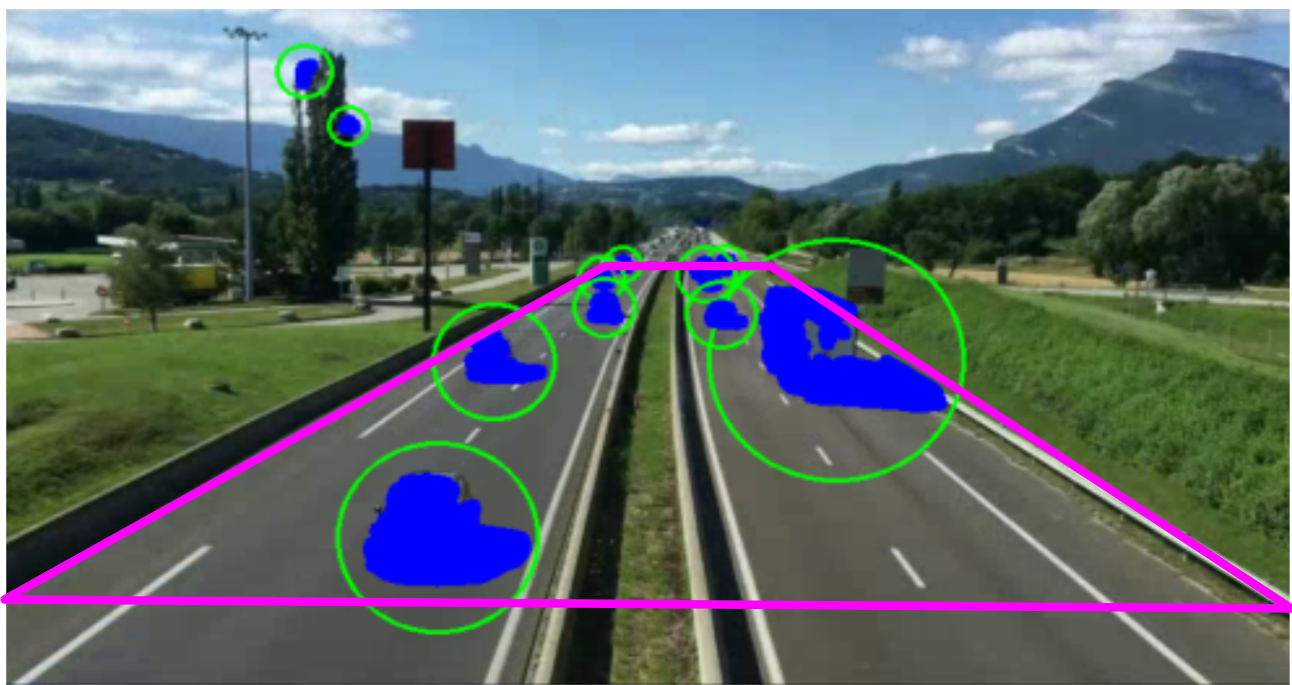}
    \vspace{5pt}
    \caption{Bounding the region of interest to focus DBSCAN detection exclusively on vehicle clusters, reducing noise from non-relevant moving objects.}
    \label{fig:bounding_area}
    \vspace{-2mm}
\end{figure}

\paragraph{Handling False Positives:}
YOLOv5 exhibited false positives, such as incorrectly identifying hoardings as vehicles, as shown in Figure \ref{fig:detection_comparison}. Although the proposed method also encountered false positives, such as identifying wind-swaying trees as potential groups, these errors can be mitigated by implementing a strategy that considers only vehicles within a predefined boundary in the camera frame, as illustrated in Figure \ref{fig:bounding_area}.

\begin{figure}[h]
    \centering
    \includegraphics[width=1\textwidth]{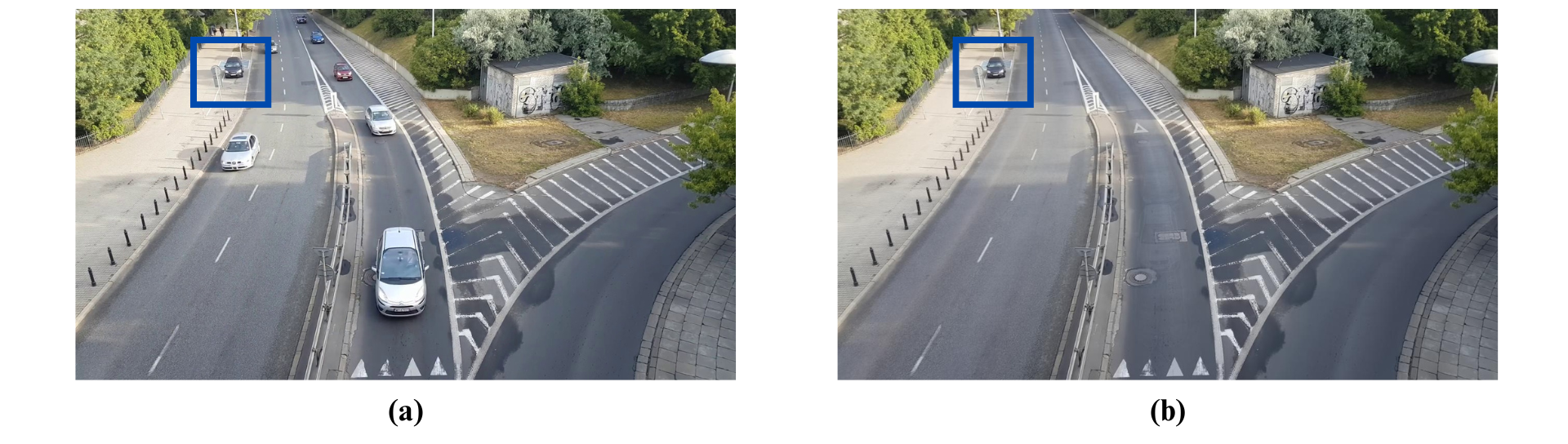}
    \caption{\textbf{(a)} Original frame for comparison, and \textbf{(b)} Parked vehicle considered as part of the modeled background.}
    \label{fig:parked_vehicle}
    \vspace{-2mm}
\end{figure}

An important observation is that stationary objects, such as vehicles parked on the road, commonly seen in developing countries, are treated as part of the background in the proposed approach, as illustrated in Figure \ref{fig:parked_vehicle}. Consequently, these objects are excluded from the foreground after differencing operation $\mathcal{D}$, providing an accurate measure of moving traffic. In contrast, directly applying the YOLOv5 detection algorithm to video frames includes parked vehicles in its count, resulting in inflated traffic measurements.

\begin{table}[ht]
\centering
\small
\captionsetup{
  font=it,               
  labelfont=bf,          
  justification=centering, 
  format=hang            
}

\caption{A runtime comparison of background extraction and object detection methods.\\ \small Note: ``-'' indicates that the method does not perform this step.}

\begin{tabular}{l c c c c c}
\toprule
Method                      & Background Reconstruction (seconds) & Detection (seconds)     \\ \midrule 
Proposed Method             &  4.5                            &       0.15              \\
K-Means Clustering          &  277                            &       -                 \\ 
YOLOv5 Object Detection     &  -                              &       0.35             \\
\bottomrule
\end{tabular}
\label{tbl:runtimes}
\end{table}

\paragraph{Evaluation of Runtime:}
We compared the runtime of the background reconstruction and detection algorithms. Background reconstruction was evaluated until a clear and refined background, which was difficult to enhance further, was achieved. The proposed algorithm used 670 frames for averaging, while the K-means approach utilized 20 frames, consistent with its original methodology. Despite the difference in frame count, the proposed method produced a clearer background with reduced runtime. In the detection phase, the DBSCAN algorithm demonstrated greater efficiency than YOLOv5. The runtimes, measured in seconds, are presented in the Table. \ref{tbl:runtimes}

\section{Conclusion}



This study presents a two-stage method for detecting moving vehicles in video frames by combining background reconstruction and clustering techniques. In the first stage, the background is reconstructed, and the foreground is isolated using the differencing operation $\mathcal{D}$. Image processing techniques are then used to refine the foreground, enabling precise object delineation. Benchmarking demonstrated that the proposed method produced a clearer background compared to K-means clustering while being approximately 61.56 times faster, based on runtime.

In the second stage, the DBSCAN algorithm is applied to identify and count vehicles, achieving accurate results through hyperparameter tuning and the exclusion of outliers. Further benchmarking revealed that DBSCAN was 2.31 times faster than YOLOv5, based on runtime comparisons.

These results highlight the method’s robustness, computational efficiency, and superior runtime performance, making it highly suitable for deployment in dynamic and congested traffic scenarios. While the results are highly promising, several challenges remain, as highlighted in the Future Work section. 
\section{Future Work}
\label{sec:future_work}




To advance this project toward real-world deployment, our immediate focus is on collecting and analyzing diverse traffic video datasets from multiple urban environments. These datasets will form the basis for validating our method under varied conditions, such as different traffic densities, lighting changes, and occlusions. 

Additionally, we plan to integrate adaptive mechanisms to handle environmental variability, including rain, fog, and shadows, which are common in urban settings. These enhancements will allow the system to auto-calibrate parameters like detection thresholds and tracking sensitivity, reducing the need for manual intervention.

We also aim to explore alternative vehicle detection approaches, including deep learning-based counting methods, and compare their performance against our current method. Moreover, the system will be designed for real-time operation, scalability across multiple traffic streams, and computational efficiency, ensuring practical deployment even on limited hardware.


\begin{thebibliography}{}

\bibitem[Busarin et~al., 2017]{7904969}
Busarin Eamthanakul, Mahasak Ketcham, and Narumol Chumuang. (2017).
The traffic congestion investigating system by image processing from CCTV camera.
In \textit{2017 International Conference on Digital Arts, Media and Technology (ICDAMT)}, pages 240–245.
DOI: 10.1109/ICDAMT.2017.7904969.

\bibitem[Escobar, 2020]{video_background_removal}
Escobar, F. (2020).
Video-Background-Removal. 
Available at: \url{https://github.com/fescobar96/Video-Background-Removal}.
Accessed: Dec. 20, 2024.

\bibitem[Gonzalez and Woods, 1993]{gonzalez1993digital}
Gonzalez, R.~C. and Woods, R.~E. (1993).
\textit{Digital Image Processing}.
Addison-Wesley Publishing Company.

\bibitem[Hahsler et~al., 2019]{dbscanArticle}
Hahsler, M., Piekenbrock, M., and Doran, D. (2019).
dbscan: Fast density-based clustering with R.
\textit{Journal of Statistical Software}, 91(1):1–30.
DOI: 10.18637/jss.v091.i01.

\bibitem[Hasan et~al., 2014]{6850751}
Hasan, M.~M., Saha, G., Hoque, A., and Majumder, M.~B. (2014).
Smart traffic control system with application of image processing techniques.
In \textit{2014 International Conference on Informatics, Electronics \& Vision (ICIEV)}, pages 1–4.
DOI: 10.1109/ICIEV.2014.6850751.

\bibitem[Hayashi and Sugimoto, 1999]{821198}
Hayashi, K. and Sugimoto, M. (1999).
Signal control system (MODERATO) in Japan.
In \textit{Proceedings 199 IEEE/IEEJ/JSAI International Conference on Intelligent Transportation Systems (Cat. No.99TH8383)}, pages 988–992.
DOI: 10.1109/ITSC.1999.821198.

\bibitem[Jocher et~al., 2020]{yolov5}
Jocher, G. and others. (2020).
YOLOv5. 
Available at: \url{https://github.com/ultralytics/yolov5}.
Accessed: Dec. 20, 2024.

\bibitem[MacQueen, 1967]{kmeans_original}
MacQueen, J. (1967).
Some methods for classification and analysis of multivariate observations.
\textit{Proceedings of the Fifth Berkeley Symposium on Mathematical Statistics and Probability}, 1:281–297.
University of California Press.

\bibitem[Mandellos et~al., 2011]{MANDELLOS20111619}
Mandellos, N.~A., Keramitsoglou, I., and Kiranoudis, C.~T. (2011).
A background subtraction algorithm for detecting and tracking vehicles.
\textit{Expert Systems with Applications}, 38:1619–1631.
DOI: 10.1016/j.eswa.2010.07.083.

\end{thebibliography}

\end{document}